# A comparative study of different feature sets for recognition of handwritten Arabic numerals using a Multi Layer Perceptron

N. Das*, A. F. Mollah[+], R. Sarkar**, S. Basu*
*Computer Science and Engineering Dept., Jadavpur University, Kolkata-70032
[+]Atrenta (I) Pvt. Ltd, Noida, Delhi
**Computer Sc. & Engg. Dept., MCKV Institute of Engineering, Liluah, Howrah-711204

**Abstract:** The work presents a comparative assessment of seven different feature sets for recognition of handwritten Arabic numerals using a Multi Layer Perceptron (MLP) based classifier. The seven feature sets employed here consist of shadow features, octant centroids, longest runs, angular distances, effective spans, dynamic centers of gravity, and some of their combinations. On experimentation with a database of 3000 samples, the maximum recognition rate of 95.80% is observed with both of two separate combinations of features. One of these combinations consists of shadow and centriod features, i. e. 88 features in all, and the other shadow, centroid and longest run features, i. e. 124 features in all. Out of these two, the former combination having a smaller number of features is finally considered effective for applications related to Optical Character Recognition (OCR) of handwritten Arabic numerals. The work can also be extended to include OCR of handwritten characters of Arabic alphabet.

**1. Introduction:**

Handwritten numeral recognition is in general a benchmark problem of Pattern Recognition and Artificial Intelligence. It is also a problem of wide commercial importance. It has applications in OCR systems, automatic pin code recognition, cheque reading, collecting data from filled in forms and so on. Compared to the problem of printed character recognition, the problem of handwritten character recognition is compounded due to variations in shapes and sizes of handwritten characters. How best these problems can be dealt with depends on discriminating     power of the feature set chosen to represent patterns of different digit classes, and the learning and generalization abilities of the pattern classifier. Considering all these, suitability of six different features sets and some of their combinations are studied under the present work for recognition of handwritten Arabic numerals with an MLP based classifier. Fig 1 shows sample images of first 10 natural numbers in Arabic.

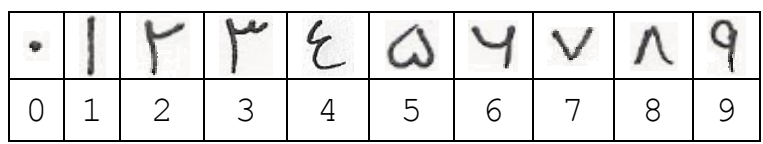

**Fig. 1.** The decimal digit set of Arabic script

In dealing with the problems of recognizing numeric patterns of varying shapes and sizes, selection of proper feature set is one of the most important factors to achieve high recognition performance. Our own interest is to design a suitably powerful feature set for recognition of handwritten Arabic numerals. So, before, a pattern recognizer can be properly designed, however, it is necessary to consider the feature extraction and data reduction problems. Any object or pattern, which can be recognized and classified, possesses a number of discriminatory properties of features. More preciously, features are the measured descriptive of a pattern which characterizes the membership of a pattern in a certain class. The task of feature extraction is to reduce the data by measuring "features" or "properties" that distinguish between different characters. These features are then passed to a classifier that evaluates the evidence presented and makes a final decision. Feature selection and extraction plays an important role in pattern recognition.

The first step in any recognition process is to consider the problem of what discriminatory features to select and how to extract these features. It is evident that the investigation of feature extraction methods has gained considerable attention since a discriminative feature set is considered the most important factor in achieving high recognition performance. Trier et al [5] present an interesting survey of feature extraction methods for off-line recognition of segmented characters. The authors describe important aspects that must be considered before selecting a specific feature extraction method. In the same direction, an interesting review of shape analysis techniques is presented by Loncaric (1998). However, discriminatory features are not easily measurable. Investigative experimentation is necessary for identifying discriminatory features before the design of pattern classifier.

## 2. FEATURE SELECTION

### 2.1 Shadow Features
Shadows features are computed by considering the lengths of projections of the digit images, as shown in Fig. 2, on the four sides and eight octant dividing sides of the minimal bounding boxes enclosing the same. Considering the lengths of projections on three sides of each such octant, 24 shadow features are extracted from each digit image, which is divided into eight octants inside the minimal box. Each value of the shadow feature so computed is to be normalized by dividing it with the maximum possible length of the projections on the respective side.

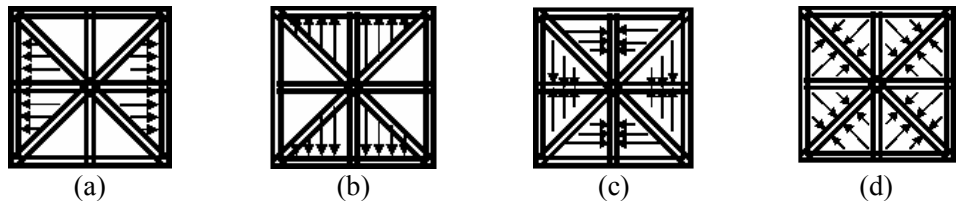

(a)          (b)          (c)          (d)

Fig 2. An illustration for shadow features.
(a-d) Direction of fictitious light rays as assume for taking the projection of an image segment on each side of all octants.

### 2.2. Octant Centroid Features
Coordinates of centroids of black pixels in all the 8 octants of a digit image are considered to add 16 features in all to the feature set. Fig. 3 shows approximate locations of all such centroids on an Arabic numeral.

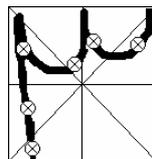

Fig 2. Centroid features of an Arabic numeral.

### 2.3 Angular Distance
To find out the angular distance of the pattern we proceed in four different ways making angles 0, 22.5, 45, 67.5 degrees respectively from each corner. An individual distance is considered as a feature. Thus we get 4 distance features for each corner and altogether 4x4 or 16 features.

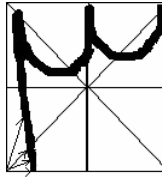

Fig. 3  Angular Distance

### 2.4 Effective Span

To calculate effective span we move from both left and right and stop whenever any black pixel is faced. This is illustrated in Fig. 4.

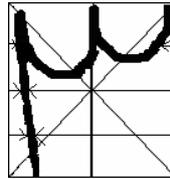

Fig. 4.  Effective Span

### 2.5  Dynamic Centre of Gravity

The centre of gravity (c. g.) will not be found out in a particular fixed way. The pattern will be splitted into 4 small patterns by the c. g. of the main pattern. Similarly each small pattern will be splitted into 4 smaller patterns. Thus we get 4 small patterns and 4x4 or 16 smaller patterns giving altogether 20 patterns. The centers of gravity of this 20 patterns give us 2x20 or 40 features. This feature is known as Dynamic Center of Gravity or simply DCG. This is illustrated in Fig. .5.

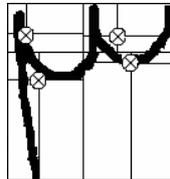

Fig. 5. Dynamic Center of Gravity

### 2.6  Longest Run Features.

 For computing longest-run features from a character image, the minimal bounding box enclosing the image is divided into 9 overlapping rectangular regions. Coordinates (r, c) of top left corners of all these regions, in terms of the row number r and the column number c, are given as follows: {(r, c) | r=0, h/4, 2h/4 and c=0, w/4, 2w/4}, where h and w denote the height and the width of the minimal bounding box respectively.  In each such rectangular region, 4 longest-run features are computed row wise, column wise and along two of its major diagonals.

The row wise longest-run feature is computed by considering the sum of the lengths of the longest bars that fit consecutive black pixels along each of all the rows of a rectangular region, as illustrated in Fig. 6(a-b). In fitting a bar with a number of consecutive black pixels within a rectangular region, the bar may be extended beyond the boundary of the region if the same is continued there. The three other longest-run features are computed in the same way but along the column wise and two major diagonal wise directions within the rectangle separately. Thus, in all, 9x4=36 longest-run features are computed from each character image. Each of these feature values is to be normalized by dividing the same with h x w. The product, h x w, represents the sum of the lengths of the bars, that fit consecutive black pixels individually in each of the four directions within the minimal square completely filled with black pixels.

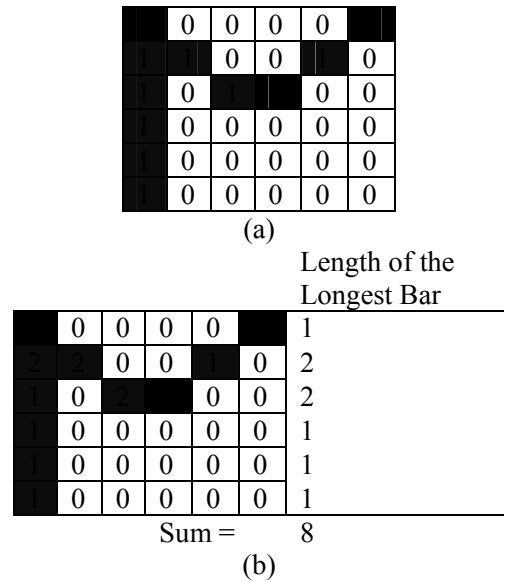

Fig 6. An illustration for computation of the row wise longest–run feature.
(a) The portion of a binary image enclosed within a rectangular region.
(b) Every pixel position in each row of the image is marked with the length of the longest bar that fits consecutive black pixels along the same row.

**3. The Feature sets:**

With the features described so far, 8 feature sets area formed separately to study their effectiveness for recognition of handwritten Arabic numerals. The feature sets are described below.

| Feature Set | Features |
|---|---|
| Set#1 | Shadow (72), octant Centroids (16) |
| Set#2 | Angular Distance(16) and Effective Span(8) |
| Set#3 | Dynamic centre of Gravity(40) |
| Set#4 | Longest Run(36) |
| Set#5 | Effective span(128) |
| Set#6 | Longest Run(36),Dynamic centre of gravity(40), Angular Distance(16), Effective span(8) |
| Set#7 | Shadow(72), Octant Centroids(16), Longest Run(36) |

The parenthesized number appearing after each feature name above represents the total number of the corresponding features.

## 4. The MLP Classifier

In the present work, an MLP classifier is employed for recognition of unknown digit patterns using the seven different feature sets. The MLP is a special kind of Artificial Neural Network (ANN). ANNs are developed to replicate *learning* and *generalization* abilities of human's behaviour with an attempt to model the functions of *biological neural networks* of the human brain.

Nowadays MLP is the mostly used classifier in the field of handwritten character recognition according to the researchers. Architecturally, an MLP is a feed-forward layered network of *artificial neurons*. Each artificial neuron in the MLP computes a *sigmoid function* of the weighted sum of all its inputs. An MLP consists of one *input layer*, one *output layer* and a number of *hidden* or intermediate *layers*, as shown in Fig 4. The output from every neuron in a layer of the MLP is connected to all inputs of each neuron in the immediate next layer of the same. Neurons in the input layer of the MLP are all basically dummy neurons as they are used simply to pass on the input to the next layer just by computing an identity function each.

The numbers of neurons in the input and the output layers of an MLP are chosen depending on the problem to be solved. The number of neurons in other layers and the number of layers in the MLP are all determined by a trial and error method at the time of its *training*. An ANN requires training to learn an unknown input-output relationship to solve a problem.

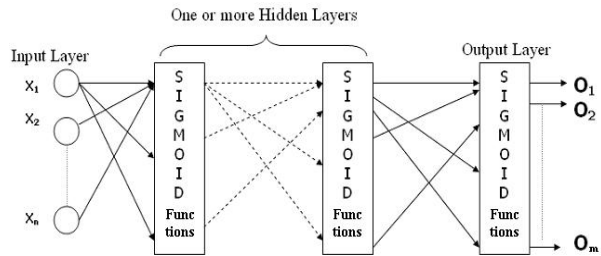

**Fig. 7.** A block diagram of an MLP shown as a feed forward neural network.

Depending on the models of ANNs, training is performed either under supervision of some teacher (i.e., with labeled data of known input-output responses) or without supervision. The MLP to be used for the present work requires supervised training. During training of an MLP *weights* or strengths of neuron-to-neuron connections, also called *synapses*, are iteratively tuned so that it can respond appropriately to all training data and also to other data, not considered at the time of training. Learning and generalization abilities of an ANN is determined on the basis of how best it can respond under these two respective situations.

The MLP classifier designed for the present work is trained with the Back Propagation (BP) algorithm. It minimizes the *sum of the squared errors* for the training samples by conducting a *gradient descent* search in the *weight space*. The number neurons in a hidden layer in the same are also adjusted during its training.

The problem of *pattern classification* involves two successive transformations as follows:

$$M \rightarrow F \rightarrow D$$

Where, M, F and D stand for the measurement space, the feature space and the decision space respectively. Once a feature set is fixed up, it is left with the design of a mapping (δ) as follows:

$$\delta: F \rightarrow D$$

ANNs with their learning and generalization abilities can approximate a general class of functions given below.

$$f: \mathbb{R}^n \rightarrow \mathbb{R}$$

Pattern classification with ANNs requires approximating δ as a *discrete valued function* shown below.

$$\delta : \mathbb{R}^n \rightarrow \{1, 2, ..m\}$$

where, n and m denotes the number of features and the number of pattern classes respectively. So an ANN based pattern classifier requires n number of neurons in the input layer and m number of neurons in the output layer. Conventionally 1-out-of-m representation is used for its output.

## 5. EXPERIMENTAL RESULTS

For preparation of the training and the test sets of samples, a database of 3000 samples is formed by collecting optically scanned handwritten Arabic specimens of first 10 natural numbers from each of 300 people of different age groups and literacy levels. A training set of 2000 samples and a test set of 1000 samples are then formed through random selection of digit samples of all classes in equal numbers from the initial database. All these samples are scaled to 32x32 pixel images first and then converted to binary images through *thresholding.*

With the pair of training and test sets, so prepared, a 2 layer MLP is designed to recognize handwritten Arabic numerals on the basis of each of seven feature sets described before. To ensure an optimal performance for each such MLP, several runs of Back Propagation (BP) algorithm with learning rate($\eta$)=0.8 and momentum term ($\alpha$)= 0.7 area executed by varying the number of neurons in the hidden layer until the number giving an optimal recognition performance is identified. For this a curve showing the variation of the recognition performance on the test set with increasing numbers of hidden layer neurons may be drawn.

The following table shows the descriptions of MLPs with their optimal recognition performances for seven different feature sets designed for this work. It can be observed from the table that an optimal recognition rate of 95.80 percent is achieved for both of feature set#1 and feature set#7. The MLP designed for the feature set#1 consists of 88 input neurons, 54 hidden layer neurons and 10 output neurons. The description of the MLP is represented as 88-54-10 in the table. The other MLP designed on the basis of feature set#7 is of the configuration 124-80-10. Though it gives the same recognition performance as the one designed with feature set#1, it requires larger number of hidden layer neurons. Considering this, the MLP (88-54-10)

Designed with feature set#1 is finally selected to ensure the optimal recognition performance out of seven feature sets considered here. The curve showing variation of the recognition performance of the MLP (on test set) with increasing numbers of hidden layer neurons is shown in Fig 8. It can be observed from the Fig.8 that the MLP shows the optimal recognition performance with 54 hidden layer neurons. This is how it was designed.

To conclude about the work, it can be stated that for recognition of handwritten Arabic numerals the feature set#1, consisting of shadow and centroid features, has produced the optimal performance compared to six other feature sets designed under the same work. An MLP classifier (88-54-10) is designed on the basis of this. The percent recognition rate of 95.8, shown by the classifier, is more or less satisfactory in respect to results achieved by others [1,2,3].

Table 1

| Sl. | Name of Feature | MLP | Recognition Rate (%) |
|---|---|---|---|
| 1 | Set#1 | **88-54-10** | **95.80** |
| 2 | Set#2 | 24-14-10 | 89.80 |
| 3 | Set#3 | 40-24-10 | 91.30 |
| 4 | Set#4 | 36-24-10 | 92.60 |
| 5 | Set#5 | 128-80-10 | 95.60 |

| 6 | Set#6 | 100-65-10 | 95.60 |
| 7 | Set#7 | 124-80-10 | 95.80 |

| No. of Hidden Neurons | % Recognition Rate |
| --- | --- |
| 20 | 94.90 |
| 25 | 95.10 |
| 30 | 95.00 |
| 33 | 95.20 |
| 35 | 95.30 |
| 40 | 95.40 |
| 45 | 95.40 |
| 50 | 95.30 |
| 54 | **95.80** |
| 55 | 95.60 |
| 60 | 95.30 |
| 65 | 95.30 |
| 70 | 95.30 |

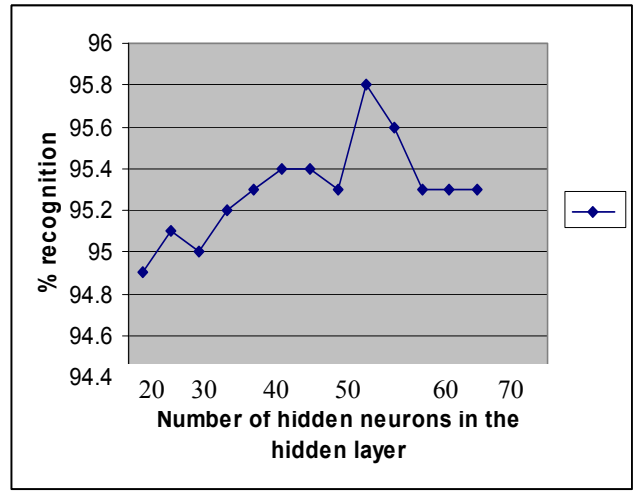

Fig. 8

In view of all this, the work can be of use in applications related to OCR of handwritten Arabic digits. It can be further extended to recognize handwritten characters of Arabic alphabet by designing a more powerful feature set in future.

**Acknowledgements**